\documentclass{article}

\usepackage{arxiv}

\usepackage[utf8]{inputenc} 
\usepackage[T1]{fontenc}    
\usepackage{hyperref}       
\usepackage{url}            
\usepackage{booktabs}       
\usepackage{amsfonts}       
\usepackage{algorithm}
\usepackage{algpseudocode}
\usepackage{hyperref}
\usepackage{amsmath}
\usepackage[textsize=tiny]{todonotes}
\usepackage{graphicx}
\usepackage{boldline}
\usepackage{xcolor}
\usepackage{xspace}
\usepackage{soul}
\usepackage[export]{adjustbox}
\usepackage[caption=false]{subfig}
\usepackage{nicefrac}       
\usepackage{microtype}      
\usepackage{lipsum}
\usepackage{rotating,tabularx}

\title{A Hybrid Evolutionary Algorithm Framework for Optimising Power Take Off and Placements of Wave Energy Converters}

\author{
  Mehdi Neshat \\
  Optimization and Logistics Group\\
  School of Computer Science\\
  The University of Adelaide\\
   Australia \\
  \texttt{mehdi.neshat@adelaide.edu.au} \\
   \And
 Bradley Alexander \\
  Optimization and Logistics Group\\
  School of Computer Science\\
  The University of Adelaide\\
   Australia \\
  \texttt{bradley.alexander@adelaide.edu.au} \\
   \And
 Nataliia Y. ~Sergiienko \\
  School of Mechanical Engineering\\
  The University of Adelaide\\
   Australia \\
  \texttt{nataliia.sergiienko@adelaide.edu.au} \\
   \And
Markus Wagner\\
  Optimization and Logistics Group\\
  School of Computer Science\\
  The University of Adelaide\\
   Australia \\
  \texttt{markus.wagner@adelaide.edu.au} \\
}

\begin{document}
\maketitle

\begin{abstract}
Ocean wave energy is a source of renewable energy that has gained much attention for its potential to contribute significantly to meeting the global energy demand.
In this research, we investigate the problem of 
maximising the energy delivered by farms of wave energy converters (WEC's). We consider state-of-the-art fully submerged three-tether converters 
deployed in arrays. The goal of this work is to use heuristic search to optimise the power output of arrays in a size-constrained environment by configuring WEC locations and the power-take-off (PTO) settings for each WEC. 
Modelling the complex hydrodynamic interactions in wave farms is expensive, which constrains search to only a few thousand model evaluations.
We explore a variety of heuristic approaches including cooperative and hybrid methods. The effectiveness of these approaches is assessed in two real wave scenarios (Sydney and Perth) with farms of two different scales. 
We find that a combination of symmetric local search with Nelder-Mead Simplex direct search combined with a back-tracking optimization strategy is able to outperform previously defined search techniques by up to 3\%.
\end{abstract}

\keywords{Renewable energy\and Evolutionary Algorithms\and Position Optimisation\and Power Take Off system\and Wave Energy Converters}

\section{Introduction}
Environmental concerns and declining costs are favouring the widespread deployment of renewable electricity generation.
Wave energy converters (WECs), in particular, offer strong potential for growth because of their high capacity factors and energy densities compared to other renewable energy technologies~\cite{drew2009review}. However, WECs are relatively new technology, which presents design challenges in the development of individual converters  and in the configuration of farms consisting of arrays of WECs. 
The  WEC model considered in this research is similar to a new generation of CETO systems that were introduced and developed by the Carnegie Clean Energy company \cite{mann2007ceto}. The CETO system is composed of an array of fully submerged three-tether converters (buoys) \cite{mann2011application}. 
The aim of this research is to maximise the absorbed power of an array (farm) of these buoys. In maximising the power produced by such an array the key factors are~\cite{de2014factors}:  (1) the layout of WECs in the sea, (2) the  power-takeoff (PTO) parameters for each WEC, (3) wave climate (wave frequencies and directions) of a specific test site, and (4) the number of WECs.

The combined search space for optimising WECs placements and PTO settings is non-linear and multi-modal. Furthermore, because of complicated and extensive hydrodynamic interactions among generators, the evaluation of each farm configuration is expensive, taking several minutes in larger farms. These factors make the use of smart and specialised meta-heuristics attractive for this problem. 

\begin{figure}[t]
\centering
\subfloat[]{
\includegraphics[clip,width=0.7\columnwidth]{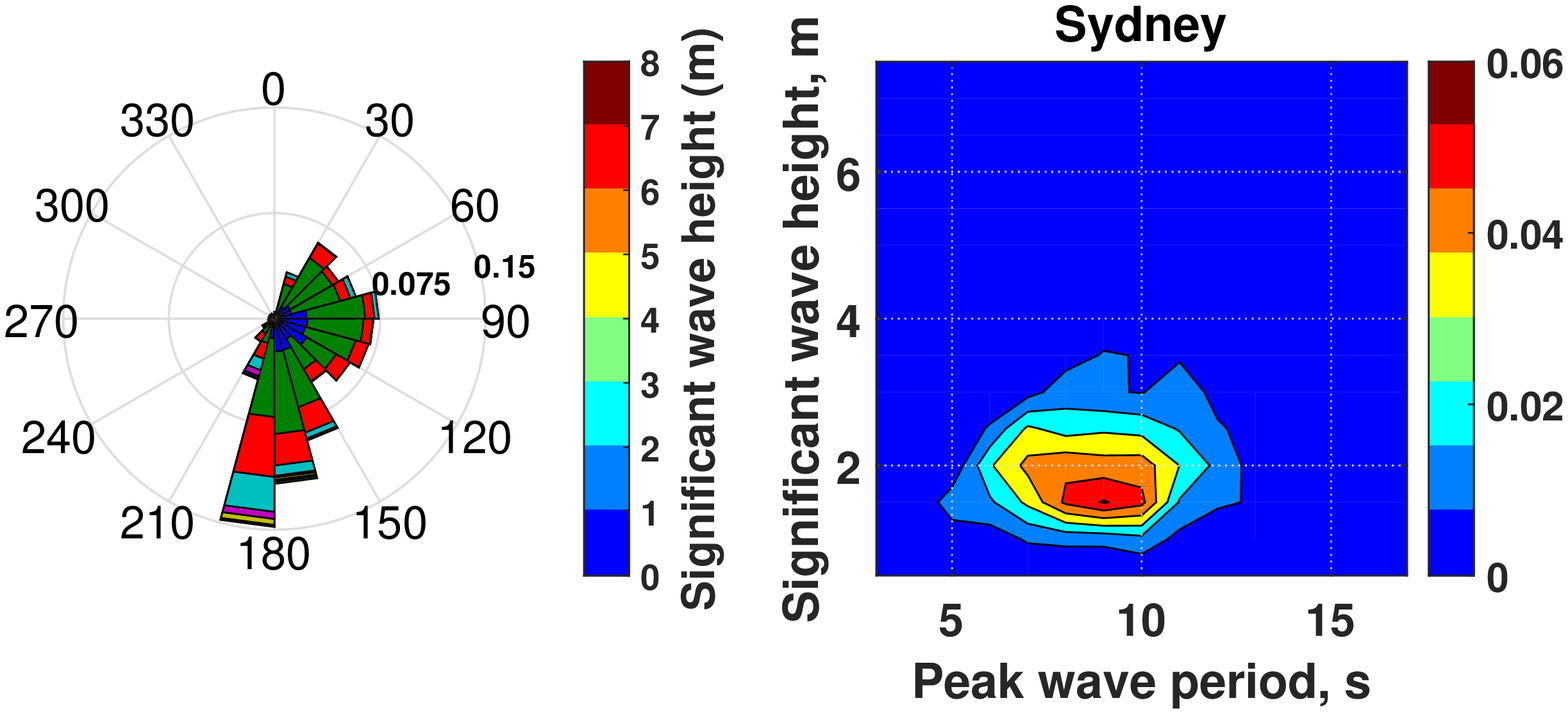}}\\
\subfloat[]{
\includegraphics[clip,width=0.7\columnwidth]{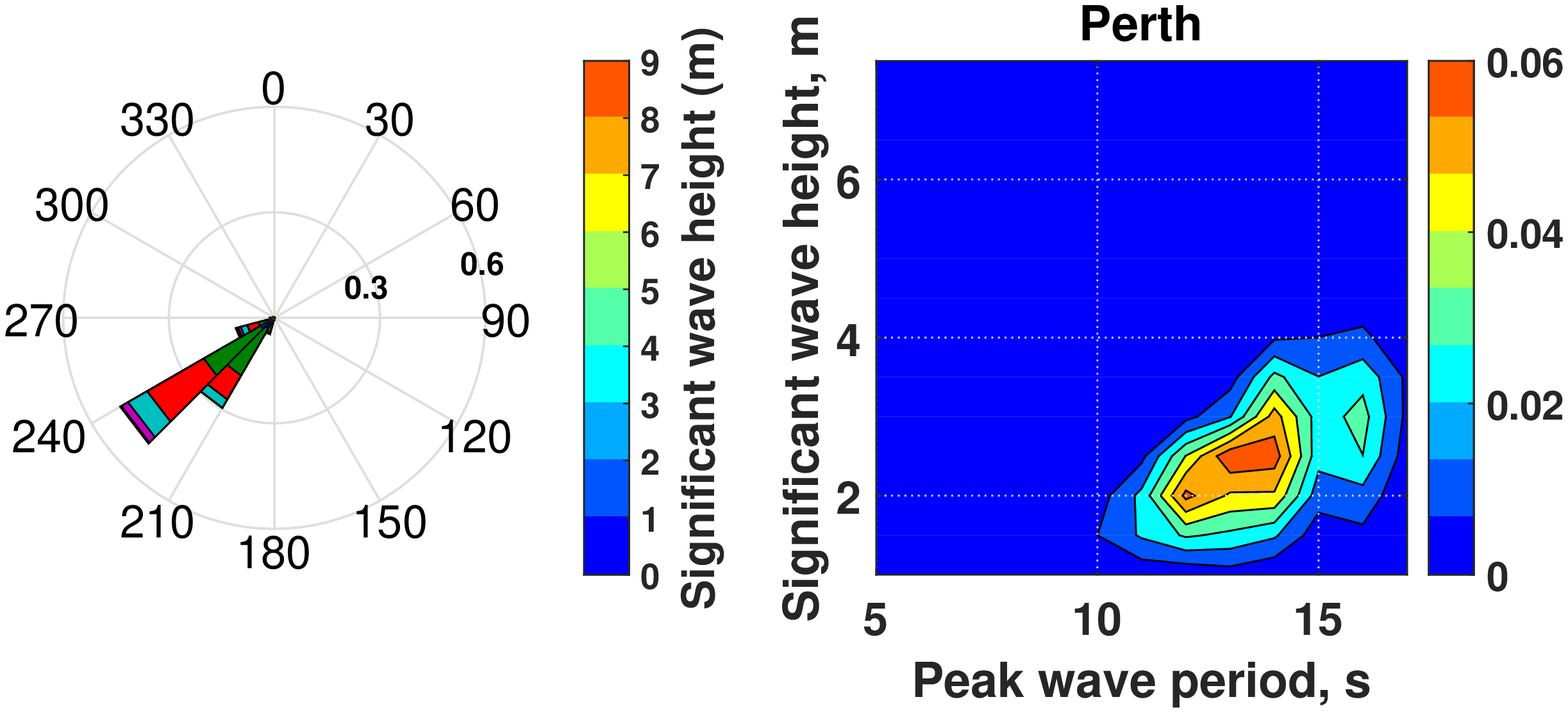}}
\caption{Wave data for two test sites in Australia: (a)~Sydney and (b)~Perth. These are: the directional wave rose (left) and wave scatter diagram (right).}%
\label{fig:wave_direct}%
\end{figure}
 
One early work \cite{child2010optimal}  used a simple uni-directional wave model to compare a custom GA with an iterative Parabolic Intersection (PI) method for placing 5 buoys. Both of these search methods deployed a high number of evaluations (37000).  
A recent study by Ruiz et al. \cite{ruiz2017layout} used another simple wave model to compare a specialised GA, CMA-ES \cite{hansen2006cma}, and glow-worm optimisation \cite{krishnanand2009glowworm} in placing buoys at positions in a discrete grid. The study found that CMA-ES converged faster than the other two methods, but ultimately produced poorer-performing layouts. 
In other recent work, Wu et al. \cite{wu2016fast} studied two EAs: a 1+1EA and CMA-ES for optimising buoy's positions in an array of fully submerged three-tether WECs using a simplified uni-directional irregular wave model. That work found that the 1+1EA with a simple mutation operator performed better than CMA-ES. More recently, Neshat et al.~\cite{neshat2018detailed} applied a more detailed wave scenario (seven wave directions and 50 wave frequencies) to evaluate a wide range of generic and custom EAs for the buoy placement. This study found that a hybrid approach (local search + Nelder-Mead) achieved better 4 and 16-buoy arrangements in terms of power produced. However, the model used by that work still embedded an artificial wave scenario. Moreover, the optimisation did not attempt to tune buoy PTO parameters to maximise the power produced by each buoy.
The optimisation of PTO parameters presents another dimension for WEC farm optimisation. PTO parameters control how WECs oscillate with the frequency of incoming waves. Maximum efficiency is achieved when converters resonate with the sea waves. However, maintaining a resonant condition is not easy because real sea waves consist of multiple different frequencies \cite{falnes2002ocean}.  
In work optimising the PTO damping of one converter (CETO 6), Ding et al. \cite {ding2016sea} applied the maximum power point tracking (MPPT) control method which is a simple gradient-ascent algorithm for the online-optimisation of the deployed WEC. The results show that the MPPT damping controller can be more effective and robust than a fixed-damping system. However, when the buoy number is increased the optimisation process becomes more complicated because of the hydrodynamic interactions between buoys. In later work Abdelkhalik et al. \cite{abdelkhalik2018optimization} used a version of the hidden genes genetic algorithm (HGGA) to control PTO parameters. While this work raised the effective energy harvested the algorithm was not compared to other methods.

In this paper, we develop a new hybrid Evolutionary framework for simultaneously optimising both placement and PTO parameters of a wave farm. We study a broad range of meta-heuristic approaches: (1) five well-known off-the-shelf EAs, (2) four alternating optimisation ideas, and (3) three hybrid optimisation algorithms.  Additionally, two new real wave scenarios from the southern coast of Australia (Perth and Sydney) with a high granularity of wave direction is used (Figure \ref{fig:wave_direct}) to evaluate and compare the performance of the proposed methods. According to our optimisation results, a new hybrid search heuristic combining symmetric local search with Nelder-Mead simplex direct search, coupled with a backtracking strategy outperforms other proposed optimisation methods in terms of the power output and computational time. 

The rest of this paper is arranged as follows. Section~\ref{sec:model} formulates the WEC model. Section~\ref{sec:opt} gives the details of the optimisation problem. The search methods are explained in Section~\ref{sec:method} and a brief characterisation of the fitness landscape is given. We present our comparative studies and experimental results in Section~\ref{sec:experiments}. Finally, Section~\ref{sec:conc} concludes this paper.

\section{Model for wave energy converters} \label{sec:model}
In this paper, we consider a fully submerged three-tether buoy model with each tether fastened to a converter installed on the seabed. We assume an optimal tether angle of 55 degrees, which was previously observed to maximise the extraction of energy from heave and surge motions~\cite{scruggs2013optimal}. Other features of the wave energy converters (WECs) used in this investigation, such as physical dimensions and submergence depth, can be found in~\cite{neshat2018detailed}.

\subsection{Power Model}
In the WEC model used here, linear wave theory is used to calculate the system dynamics~\cite{ sergiienko2016frequency}. This model includes three different key forces:
\begin{enumerate}
\item The wave excitation force ($F_{exc,p}(t)$) combines the incident and diffracted waves forces from generators in a fixed location. 
\item The radiation force ($F_{rad,p}(t)$), derived by the oscillating body due to their motion independent of  incident waves.
\item Power take-off (PTO) force ($F_{pto,p}(t)$) is the control force applied to the buoy from the PTO machinery. 
\end{enumerate}

Through these forces, the buoys can affect each other's output through hydrodynamic interactions. The complex nature of these interactions, which can either be constructive or destructive, makes the calculation of farm layout and PTO parameter settings a challenging optimisation problem.
The dynamic equation that describes a buoy motion in ocean waves has the form:
\begin{equation}\label{eqn-power}
M_p\ddot{X}_p(t)=F_{exc,p}(t)+F_{rad,p}(t)+F_{pto,p}(t)
\end{equation}
\noindent where $M_p$ is the mass matrix of a $p_{th}$ buoy, ${X}_p(t)$ is the buoy displacement expressed as surge, heave and sway. Finally, the power take-off system is modeled as a linear spring-damper system. For each mooring line two control factors are involved: the damping $D_{pto}$ and stiffness $K_{pto}$ coefficients. Therefore, Equation (\ref{eqn-power}) can be written in a frequency domain for all WECs in a farm as:
\begin{equation}\label{Ex-eqn-power}
\hat{F}_{exc,\Sigma}=((M_{\Sigma}+A_{\sigma}(\omega))j\omega+B_{\sigma}(\omega)-\frac{K_{pto,\Sigma}}{\omega}j+D_{pto,\Sigma})\ddot{X}_{\Sigma}
\end{equation}
The hydrodynamic parameters ($A_{\Sigma}(\omega))$ and $B_{\Sigma}(\omega)$ ) are calculated from the semi-analytical model described in \cite{wu1995radiation}. In addition, $K_{pto,\Sigma}$ and $D_{pto,\Sigma}$ are control factors, described above, which can be adjusted to maximise the power output of each buoy. The total power output of the layout is computed by Equation~(\ref{total-power}):
\begin{equation}\label{total-power}
P_{\Sigma}= \frac{1}{4}(\hat{F^*}_{exc,\Sigma}\ddot{X}_{\Sigma}+\ddot{X^*}_{\Sigma}\hat{F}_{exc,\Sigma})-\frac{1}{2}\ddot{X^*}_{\Sigma}B\ddot{X^*}_{\Sigma}
\end{equation}
Additionally, the q-factor ($q$) of the array measures the efficiency of a entire wave farm as compared to the power output from $N$ isolated WECs.
For a given layout, the $q$-factor can be calculated as:

\begin{equation}\label{q-factor}
q=\frac{P_{\sum}}{\sum_{i=1}^{N}P_i}.
\end{equation}
$q>1$ indicates constructive interference between WECs. The main purpose of this study is maximising the total power output: $P_{\Sigma}$ for $N$ buoys within a constrained farm area.

\section{Optimisation problem formulation}\label{sec:opt}
The formulation of the optimisation problem in this paper can be declared as:
\[
  P_{\Sigma}^* = \mbox{\em argmax}_{\mathit{X,Y,K_{pto},D_{pto}}} P_{\Sigma}(\mathit{X,Y,K_{pto},D_{pto}})
\]
\noindent where $P_{\Sigma}(\mathit{X,Y,K_{pto},D_{pto}})$ is the mean power obtained by placements and PTO parameters of the buoys in a 2-D coordinate system at $x$-positions: $\mathit{X}=[x_1,\ldots,x_N]$, $y$-positions: $\mathit{Y}=[y_1,\ldots,y_N]$ and corresponding Power Take-off parameters including $\mathit{K_{pto}}=[k_1,\ldots,k_N]$ and $\mathit{D_{pto}}=[d_1,\ldots,d_N]$ . In the experiments here $N \in \{4,16\}$. 
\paragraph{Constraints}
All buoy locations $(x_i,y_i)$ are constrained to a square search space $S=[x_l,x_u]\times[y_l,y_u]$: where $x_l=y_l=0 ~and~x_u=y_u=\sqrt{N * 20000}\,m$. This allocates $20000m^2$ of farm-area per-buoy. Moreover, a safety distance for maintenance vessels must be maintained between buoys of at least 50 meters. For spring and damper coefficients the boundary constraints are $d_l=5\times10^4, d_u=4\times10^5$ and $k_l=1, k_u=5.5\times10^5$. For any array $\mathit{X,Y}$ the sum-total violations of the inter-buoy distance calculated in meters, is:

\vspace{2mm}\hspace{5mm}$\mbox{\em{Sum}}_{\mbox{\em dist}}= \sum_{i=1}^{N-1}\sum_{j=i+1}^{N} 
(\mbox{\em{dist}}((x_i,y_i),(x_j,y_j))-50), $

\hspace{38mm}$\mbox{if } \mbox{\em{dist}}((x_i,y_i),(x_j,y_j))<50$ \mbox{else 0}
\vspace{-2mm}

where $dist((x_i,y_i),(x_j,y_j))$ is the Euclidean distance between buoys $i$ and $j$. The penalty function of the power output (in Watts) is computed by $(\mbox{\em{Sum}}_{\mbox{\em{dist}}}+1)^{20}$. The penalty strongly encourages feasible buoy placements.
This penalty is also used to handle farm-boundary constraints. For the $D_{pto}$ and $K_{pto}$ parameters, we handle constraint violations by setting the parameter to the nearest valid value. 


\paragraph{Computational Resources}
In this paper, we aim to compare a various heuristic search methods, for 4 and 16 buoy arrays, in two realistic wave scenarios. We allocate a time budget for each optimization run of three days on dedicated platform with a 2.4GHz Intel 6148 processor running 12 processes in parallel with 128GB of RAM. Note, that where the search heuristic allows, we tune algorithm settings to utilise this time budget. The software environment running the function evaluations and the search algorithm is MATLAB R2017. On this platform, parallelisation provides up to 10 times speedup. 

\section{Optimisation Methods}\label{sec:method}
In this research, our search methods employ three broad strategies.  
The first strategy is to optimise all decision variables
at once. This means that for a 16-buoy farm we search in $16\times4$ dimensions simultaneously. Here, we test five heuristics that apply this strategy. The second strategy is to optimise the positions and PTO parameters of all buoys in an alternating cooperative algorithm \cite{bezdek2003convergence}. We test four different methods that apply this strategy. Finally, the third strategy, used in \cite{neshat2018detailed} 
is to place and optimise each buoy in sequence. Here, we deploy this strategy for three hybrid EAs. Details of the algorithms tested for each strategy follow.

\begin{center}
 \begin{algorithm}\small
\caption{NM+Mutation}\label{alg:NM_M}
\begin{algorithmic}[1]
\Procedure{Nelder-Mead + Mutation (all Dims)}{}\\
 \textbf{Initialization}
 \\$\mathit{size}=\sqrt{N*20000}$  \Comment{Farm size}
 \\$\mathit{S}=\{\langle x_1,y_1,k_1,d_1 \rangle,\ldots,\langle x_N,y_N,k_N,d_N \rangle\}$ \Comment{Positions\&PTOs}
\\$\mathit{bestEnergy}=0$ \Comment{Best energy so far}
\\$\mathit{bestLayout}=[\mathbf{S}]$ \Comment{Best layout so far}
\\$\mathit{EIRate}=0$ \Comment{Energy Improvement rate }\\
 \textbf{Iterative search}
\While{{\em{stillTime()}}}
 \State $(\mathit{S'},\mathit{energy})$=   {\em{NM\_Search}}$(\mathbf{S},\mathit{MaxEval})$ 
 \Comment{Local search}
 \State $\mathit{EIRate}$=   \em{ComputeEIRate}$(\mathit{energy},\mathit{bestEnergy})$ 
  \If {$\mathit{energy} > \mathit{bestEnergy}$} 
      \State $\mathit{bestEnergy}=\mathit{energy}$ \Comment{Update energy}
      \State $\mathit{bestLayout}=\mathit{S'}$ \Comment{Update layout}
      \State $\mathit{S}=\mathit{S'}$
 \EndIf
 \If {$\mathit{EIRate} = 0$} 
      \While{ $(\mathit{EIRate}= 0)$}
      \State  $\mathit{S'}=\mathit{randn}(\sigma) + \mathit{S}$ \Comment{ new buoys Position\&PTO}
       \State $\mathit{energy}= \mathit{Eval}(\mathit{S'})$ 
       \State $\mathit{EIRate}$=   \em{ComputeEIRate}$(\mathit{energy},\mathit{bestEnergy})$ 
  \EndWhile
  \If {$\mathit{energy} > \mathit{bestEnergy}$} 
      \State $\mathit{bestEnergy}=\mathit{energy}$ \Comment{Update energy}
      \State $\mathit{bestLayout}=\mathit{S'}$ \Comment{Update layout}
      \State $\mathbf{S}=\mathit{S'}$
 \EndIf
 \EndIf
 
\EndWhile
\State \textbf{return} $\mathit{bestLayout},\mathit{bestEnergy}$ \Comment{Final Layout}
\EndProcedure
\end{algorithmic}
\end{algorithm}
      \end{center}

\subsection{Evolutionary Algorithms (All-at-once)}\label{sec:allatonce}
\label{subsec:EAs}
For the first strategy, five well-known off-the-shelf EAs are deployed to simultaneously optimise all problem dimensions. 
(Positions+PTOs). 
These EAs are: (1) covariance matrix adaptation evolutionary-strategy (CMA-ES) \cite{hansen2006cma} 
with the default $\lambda=12$, for 4-buoy layouts and  and $\lambda=16$ for 16-buoy layouts; 
(2) Differential Evolution (DE)~\cite{storn1997differential},  with parameter settings of 
$\lambda=50,30$, respectively for 4 and 16-buoy layouts, and $F=0.5$, $P_{cr}=0.5$; 
(3) a (1+1)EA \cite{eiben2007parameter} that mutates buoys' location and PTO parameters with a probability of $1/N$ using a normal distribution ($\sigma=0.1\times(U_b-L_b)$); (4) Particle Swarm optimisation (PSO) \cite{eberhart1995new}, with 
$\lambda$= DE settings,~$c_1=1.5, c_2=2,\omega=1$ (linearly decreased); (5) Nelder-Mead simplex direct search (NM) \cite{lagarias1998convergence}  is combined with a mutation operator (Nelder-Mead+Mutation or NM-M). The mutation operation is applied when the NM has converged to a solution before exhausting its computational budget, so that it can explore other parts of the solution-space (Algorithm \ref{alg:NM_M}). 
 \begin{center}
 \begin{algorithm}\small
\caption{CMAES+NM}\label{alg:CMAES_NM}
\begin{algorithmic}[1]
\Procedure{(2+2)CMA-ES + Nelder-Mead (all Dims)}{}\\
 \textbf{Initialization}
 \\$\mathit{size}=\sqrt{N*20000}$  \Comment{Farm size}
 \\$\mathit{NPop}=2$  \Comment{Population size}
 \\$\mathit{S}=\{\langle x_1,y_1,k_1,d_1 \rangle,\ldots,\langle x_N,y_N,k_N,d_N \rangle\}$ \Comment{Positions\&PTOs}
 \\$\langle \mathit{S_1,S_2} \rangle=\mathit{Decompose(S)}$ \Comment{Decomposing} 
\\$\mathit{S_1}=\{\langle x_1,y_1 \rangle,\ldots,\langle x_N,y_N \rangle\}=\bot$ \Comment{Positions}
\\$\mathit{S_2}=\{\langle k_1,d_1 \rangle,\ldots,\langle k_N,d_N \rangle\}=\bot$ \Comment{PTO parameters}
\\$\mathbf{Pop}=\mathit{initPopulation(\mathbf{\{S_1,S_2\}},\mathit{NPop})}$ 
\\$\mathit{bestEnergy}=0$ \Comment{Best energy so far}
\\$\mathit{bestPosition}=[\mathit{S_1}]$ \Comment{Best Position so far}
\\$\mathit{bestPTO}=[\mathit{S_2}]$ \Comment{Best PTO parameters so far}
\\$\mathit{MaxEval}=\mathit{MaxIterC}\times \mathit{NPop}$ 
\\ \textbf{Cooperative search}
\While{{\em{stillTime()}}}
 \State \textbf{\em{Position Optimization}}
 \State $(\mathbf{Pop_{S_1}},\mathit{energies})$=   {\em{2+2CMA-ES}}$(\mathbf{Pop},\mathit{MaxIterC})$ 
 \State $\langle\mathit{bestPosition,bestIndex}\rangle$=   \em{FindBest}$(\mathbf{Pop_{S_1}},\mathit{energies})$ 
 \State \textbf{PTO Optimization}
  \State $(\mathit{bestEnergy},\mathit{bestPTO})$=   {\em{NM}}$(\mathbf{Pop(\mathit{bestIndex})},\mathit{MaxEval})$ 
  \State $\mathbf{Pop_{S_2}(\mathit{bestIndex})}= \mathit{bestPTO}$  \Comment{Update best solution}
\EndWhile
\State \textbf{return} $\mathit{bestPosition},\mathit{bestPTO},\mathit{bestEnergy}$ 
\EndProcedure
\end{algorithmic}
\end{algorithm}
     \end{center}
\subsection{Alternating optimisation methods (Cooperative ideas)}\label{sec:alternate}
Optimising both positions and PTO parameters of a WEC array simultaneously can be challenging because of the high number of dimensions and heterogeneous kinds of variables. There is a natural division of variables into two subsets which might, at least in part, be optimised separately.
In this section, we describe a  set of alternating optimisation techniques which combine one evolutionary algorithm idea such as CMA-ES, DE, and 1+1EA, with  Nelder-Mead. In addition, a cooperative, Dual-DE (DE+DE), algorithm is also described. 
The details of each are given next. 
\subsubsection{(2+2)CMA-ES + Nelder-Mead}
This alternating strategy applies CMA-ES with $\mu=\lambda=2$ for $\mathit{iter}=25$ iterations to optimise buoy positions. Then the best solution is selected and NM is applied to PTO settings for $\mathit{iter}*\lambda$ iterations. This improved setting is then given to the CMA-ES population for another round of optimisation. The CMA-ES and NM optimisation processes are alternated until the time budget expires. Algorithm \ref{alg:CMAES_NM} shows the process of the CMAES-NM approach.

\subsubsection{DE + Nelder-Mead}(DE-NM)
This method alternates DE, for buoy-positions, and NM for PTO parameters, using the same iteration settings as above until the time budget runs out.
\subsubsection{1+1EA + Nelder-Mead} (1+1EA-NM)
This method alternates a 1+1 EA, for buoy positions, and NM, for PTO parameters until the time budget runs out. The iteration settings for the 1+1EA are, respectively, 
200 and 50 times, for 4 and 16-buoy layouts.  The same limits are also used for the NM optimisation rounds.
\subsubsection{Dual-DE}
This method uses the same parameter settings as described for DE in subsection~\ref{subsec:EAs} to optimise both buoy positions and PTO parameters in parallel. After $\mathit{iter}$ iterations the improved values from the positional and PTO optimisations are exchanged.
This iterative pattern continues until the time budget runs out.   

\subsection{Hybrid optimisation algorithms }\label{sec:hybrid}
In other WEC-related research~\cite{neshat2018detailed}, it was found that applying local search around the neighborhood of previously placed buoys could help exploit constructive interactions between buoys. The following methods exploit this observation by placing and optimising the position and PTO parameters of one buoy at a time. 
\subsubsection{Local Search + Nelder-Mead(LS-NM)}
 This method places buoys sequentially. The position of each buoy placement is optimised by sampling at a normally-distributed random offset ($\sigma=70m$) from the previous buoy position. The sampled location giving the highest output is chosen. In our experiments we try three different numbers of samples: ($N_s=2^4, 2^5 ~and ~2^6$).  After the best position is selected, we optimise the PTO parameters of the last placed buoy using $N_s$ iterations of  Nelder-Mead search. This process is repeated until all buoys are placed. Note that, the $\mathit{Eval}$ function of LS-NM is parallelised on a per-wave-frequency basis. An example of 16-buoy layout that is built by LS-NM(16s) and the sampling process used to build it,  is shown in Figure \ref{fig:three_solutions}(a). The details of the proposed method can be seen in Algorithm \ref{alg:NMLS}.
      \begin{center}
\begin{algorithm}\small
\caption{$\mathit{LS}+NM$}\label{alg:NMLS}
\begin{algorithmic}[1]
\Procedure{Local Search + Nelder-Mead (2 Dims)}{}\\
 \textbf{Initialization}
 \\$\mathit{size}=\sqrt{N*20000}$  \Comment{Farm size}
\\$\mathit{S}=\{\langle x_1,y_1,k_1,d_1 \rangle,\ldots,\langle x_N,y_N,k_N,d_N \rangle\}$ \Comment{Positions\&PTOs}
 \\$\langle \mathit{S1,S2} \rangle=\mathit{Decompose(S)}$ \Comment{Decomposing} 
\\$\mathit{S1}=\{\langle x_1,y_1 \rangle,\ldots,\langle x_N,y_N \rangle\}=\bot$ \Comment{Positions}
\\$\mathit{S2}=\{\langle k_1,d_1 \rangle,\ldots,\langle k_N,d_N \rangle\}=\bot$ \Comment{PTO parameters}
\\ $\mathit{S1}_{(1)}=\{\langle size/2,0\rangle\}$  \Comment{first buoy position}
\\ $\mathit{S2}_{(1)}=\{\langle rand\times Max_{k},rand\times Max_{d}\rangle\}$ \Comment{first buoy k and d}
\\$(S2_{(1)})$={\em{NM}}$(\mathit{S1_{(1)}},S2_{(1)},MaxEN)$ {\Comment{Optimise first buoy PTO}}
\\$\mathit{bestPosition}=S1_{(1)}$; $\mathit{bestPTO}=S2_{(1)}$
 \For{ $i$ in $[2,..,N]$ }
\State $\mathit{iters}=MaxSN$ \Comment{Number of local samples}
\State$\mathit{bestEnergy}=0;$
 \State \textbf{\em{Position Optimization}}
\For{$j$ in $[1,..,\mathit{iters}]$}
  \While{ {\em{not feasible position}} }
       \State  $\mathit{tPos}=\mathit{randn}(\sigma) + \mathit{S1}_{(i-1)}$ \Comment{ new buoy position}
      
  \EndWhile
  \State $\mathit{energy}=\mathit{Eval}([S1_{(1)},\ldots,S1_{(i-1)},tPos])$
  \If{$\mathit{energy}>\mathit{bestEnergy}$}
    \State $\mathit{S1_{(i)}}=tPos$ \Comment{Update last buoy position}
    \State $\mathit{bestPosition}=[S1_{(1)},\ldots,S1_{(i-1)},S1_{(i)}]$ 
    \State $\mathit{bestEnergy}=\mathit{energy}$
    \EndIf
  \EndFor
  \State \textbf{\em{PTO Optimization}}
  \State $(S2_{(i)},\mathit{energy})$={\em{NM}}$(\mathit{bestPosition},S2_{(i-1)},MaxEN)$ 
  \If{$\mathit{energy}>\mathit{bestEnergy}$}
    \State $\mathit{bestPTO}=[S2_{(1)},\ldots,S2_{(i-1)},S2_{(i)}]$ 
    \State $\mathit{bestEnergy}=\mathit{energy}$
    \EndIf

\EndFor 
\State \textbf{return} $\mathit{bestPosition},\mathit{bestPTO},\mathit{bestEnergy}$  \Comment{Final Layout}
\EndProcedure
\end{algorithmic}
\end{algorithm}
    \end{center}
\subsubsection{Symmetric Local Search + Nelder-Mead (SLS+NM(2D))}
This method also places one buoy at a time, but performs a more systematic local search. The search starts by placing the first buoy in the middle of the bottom of the field and then uses NM to optimise the PTO parameters for $25$ iterations. 

For each subsequent buoy placement, eight local samples are made in different sectors starting at angles: $\{angles=[0, 45, 90,...,315]\}$ and bounded by a radial distance of between $50$ (safe distance) and $50+R'$. Within each sector a buoy position is sampled uniformly. Our strategy for handling infeasible solutions is that we refuse them and if all symmetric solutions are infeasible, a feasible layout is produced using uniform random sampling.

After finding the best sample among the eight local samples, two extra samples are done for increasing the resolution of the search direction. The angles of these two samples are $\pm~15^{o}$ plus the best angle sample. The candidate position is then selected from the 8 original samples plus these two extra samples based on the buoy's energy output. 

In the next step a check is done to see if the PTO optimisation process for the previously placed buoy (using NM) had a high percentage improvement in its last step. A large improvement indicates that there is scope to improve energy production, in this environment, by giving priority to PTO optimisation. Thus, if the last PTO search step for the last buoy is greater than $0.01\%$ then we optimise PTO parameters for $25$ iterations using NM. Otherwise we check to see if the last {\em{position}} optimisation converged to within $0.01\%$ and if so, we optimise position instead. Otherwise we choose between optimising PTO or position parameters for this buoy at random. 

Note that this design assigns optimisation resources to PTO parameters as a first priority because we have observed stronger gains in output from tuning PTO parameters. Position parameters are given priority only when the PTO parameters for the last buoy were observed to be close to a local optimum. 
Algorithm \ref{alg:SLSNM} describes this method in detail. In addition, experiments were run with different starting buoy positions of  were run with bottom center (C), bottom right (BR) and a uniform random position (r).
\begin{algorithm}[t]\small
\caption{$\mathit{SLS}+NM(2D)$}\label{alg:SLSNM}
\begin{algorithmic}[1]
\Procedure{Symmetric Local Search + Nelder-Mead}{}\\
 \textbf{Initialization}
 \\$\mathit{size}=\sqrt{N*20000}$  \Comment{Farm size}
  \\$\mathit{angle=\{0,45,90,\ldots,315\}}$ \Comment{symmetric samples angle}
 \\ $\mathit{iters}=Size([angle])$ \Comment{Number of symmetric samples}
\\$\mathit{S}=\{\langle x_1,y_1,k_1,d_1 \rangle,\ldots,\langle x_N,y_N,k_N,d_N \rangle\}$ \Comment{Positions\&PTOs}
 \\$\langle \mathit{S1,S2} \rangle=\mathit{Decompose(S)}$ \Comment{Decomposing} 
\\$\mathit{S1}=\{\langle x_1,y_1 \rangle,\ldots,\langle x_N,y_N \rangle\}=\bot$ \Comment{Positions}
\\$\mathit{S2}=\{\langle k_1,d_1 \rangle,\ldots,\langle k_N,d_N \rangle\}=\bot$ \Comment{PTO parameters}
\\ $\mathit{S1}_{(1)}=\{\langle size/2,0\rangle\}$  \Comment{first buoy position}
\\ $\mathit{S2}_{(1)}=\{\langle rand\times Max_{k},rand\times Max_{d}\rangle\}$ \Comment{first buoy k and d}
\\ $\mathit{energy}=\mathit{Eval}([S1_{(1)},S2_{(1)}])$
\\$(\mathit{S2_{(1)},bestEnergy})$={\em{NM}}$(\mathit{S1_{(1)}},S2_{(1)})$ \Comment{Optimise first buoy PTOs}
\\{$(\mathit{ImPTOrate})$={\em{ComputeImrate}}$(\mathit{bestEnergy,energy})$ }
\\$\mathit{bestPosition}=S1_{(1)}$; $\mathit{bestPTO}=S2_{(1)}$
\\$\mathit{ImPorate=1}$ {\Comment{optimisation improvement rate Position }}
 \For{ $i$ in $[2,..,N]$ }
\State$\mathit{bestEnergy}=0;$
 \For{$j$ in $[1,..,\mathit{iters}]$}
\State $(Sample_{j},\mathit{energy_{j}})$={\em{SymmetricSample}}$(\mathit{angle_j},S1_{(i-1)})$ 
    \If{ {\em{$Sample_j$ is feasible \& $energy_j$ > bestEnergy}} }
       \State  $\mathit{tPos}=\mathit{Sample_j}$ \Comment{ Temporary buoy position}
      \State $\mathit{bestEnergy}=\mathit{energy_j}$
      \State $\mathit{bestAngle}=\mathit{j}$
  \EndIf
  \EndFor
  \If{ {\em{No feasible solution is found}} }
      \State $(Sample_{1},\mathit{energy_{1}})$={\em{rand($\mathit{S1_{(i-1)}}$)}} 
 \EndIf
   \State $(Es_1,Es_2)$={\em{SymmetricSample}}$(\mathit{bestAngle\pm15},S1_{(i-1)})$ 
   \State $(S1_{(i)},\mathit{energy})$={\em{FindbestS}}$(\mathit{tPos},Es_1,Es_2)$ 
   \If{$\mathit{ImPTOrate\ge 0.01\%}$}
   \State \textbf{\em{PTO optimisation}}
  \State $(S2_{(i)},\mathit{energy})$={\em{NM}}$(\mathit{bestPosition},S2_{(i-1)},MaxEN)$ 
   \State{$(\mathit{ImPTOrate})$={\em{ComputeImrate}}$(\mathit{bestEnergy,energy})$ }
  \If{$\mathit{energy}>\mathit{bestEnergy}$}
    \State $\mathit{bestPTO}=[S2_{(1)},\ldots,S2_{(i-1)},S2_{(i)}]$ 
    \State $\mathit{bestEnergy}=\mathit{energy}$
    \EndIf
 \ElsIf{$\mathit{ImPorate\ge 0.01\%}$}
 \State \textbf{\em{Position optimisation}}
  \State $(S1_{(i)},\mathit{energy})$={\em{NM}}$(\mathit{S1_{(i)},bestPTO},MaxEN)$ 
   \State{$(\mathit{ImPorate})$={\em{ComputeImrate}}$(\mathit{bestEnergy,energy})$ }
  \If{$\mathit{energy}>\mathit{bestEnergy}$}
    \State $\mathit{bestPosition}=[S1_{(1)},\ldots,S1_{(i-1)},S1_{(i)}]$ 
    \State $\mathit{bestEnergy}=\mathit{energy}$
    \EndIf
  
      \Else
    \State \em{Optimise one of buoy Position or PTO randomly}
\EndIf
\EndFor 
\State \textbf{return} $\mathit{bestPosition},\mathit{bestPTO},\mathit{bestEnergy}$  \Comment{Final Layout}
\EndProcedure
\end{algorithmic}
\end{algorithm}
 \begin{algorithm} \small
\caption{$\mathit{Backtracking\, optimisation\, Algorithm\, (BOA)}$}\label{alg:B}
\begin{algorithmic}[1]
\Procedure{BOA ($\mathit{Position},\mathit{PTOs},\mathit{Energy}$ )}{}\\
 \textbf{Initialization}
  \\$\mathit{S1}=\{\langle x_1,y_1 \rangle,\ldots,\langle x_N,y_N \rangle\}=\mathit{Position}$ \Comment{Positions}
\\$\mathit{S2}=\{\langle k_1,d_1 \rangle,\ldots,\langle k_N,d_N \rangle\}=\mathit{PTOs}$ \Comment{PTO parameters}
\\ $\mathit{energy}=([E_{1},E_{2},\ldots,E_{N}])=\mathit{Energy}$ \Comment{Buoys energy}
\\ $N_w=N/4$
\\$(\mathit{WIndex})$={\em{FindWorst}}$(\mathit{energy},N_w)$ 
\Comment{Find worst buoys power}
 \For{ $i$ in $[1,..,N_w]$ }
   \State \textbf{\em{PTO optimisation}}
  \State $(S2_{\mathit{WIndex}(i)},\mathit{energy}_{\mathit{WIndex}(i)})$={\em{NM}}$(S2_{\mathit{WIndex}(i)},MaxEN)$ 
   
     \State \textbf{\em{Position optimisation}}
  \State $(S1_{\mathit{WIndex}(i)},\mathit{energy}_{\mathit{WIndex}(i)})$={\em{NM}}$(\mathit{S1_{\mathit{WIndex}(i)}},MaxEN)$ 
     \EndFor 
\State \textbf{return} $\mathit{S1},\mathit{S2},\mathit{energy}$  \Comment{Final Layout}
\EndProcedure
\end{algorithmic}
\end{algorithm}
\subsubsection{Symmetric Local Search + Nelder-Mead + Backtracking (SLS-NM-B)}
The general idea of SLS-NM-B is like SLS-NM but with two differences. The first difference is optimising the initial buoy PTO settings by Nelder-Mead and then to share this configuration with the next placed buoys for speeding up the search process and saving computational time. Therefore, after applying  symmetric local sampling and finding the best position, Nelder-Mead search tries to improve just the position (2D) of the new buoy. 

The second contribution is applying a backtracking optimisation idea (described in Algorithm~\ref{alg:B}). As the search process of SLS is based on the greedy selection, we never come back to enhance previous buoys' attributes, so introducing  backtracking  can be effective for maximising total power output. Among all placed buoys in the array, the worst $round(N\times0.25)$ buoys in terms of power are chosen and Nelder-Mead search is then used to optimise the position (2D) and PTO settings (2D) of these buoys in a bi-level optimisation process. This procedure is  called SLS-NM-B1. We can observe the performance of SLS-NM-B1 in Figure~\ref{fig:three_solutions}(b,c). This shows how the eight symmetric samples are done and the effect of the later backtracking process which refines buoy placements. 
A second version of this algorithm is proposed (SLS-NM-B2) to evaluate the effectiveness of optimising both position and PTOs of each buoy (4D) simultaneously instead of in a bi-level search. Other details of the  backtracking algorithm are the same. 

\section{Experiments}\label{sec:experiments}
This section first presents a brief landscape analysis for PTO parameters for two wave scenarios (Perth and Sydney). We then present detailed results comparing the different search heuristics outlined in the previous section. 
\begin{figure}[tb]
\centering
\includegraphics[width=0.7\textwidth]{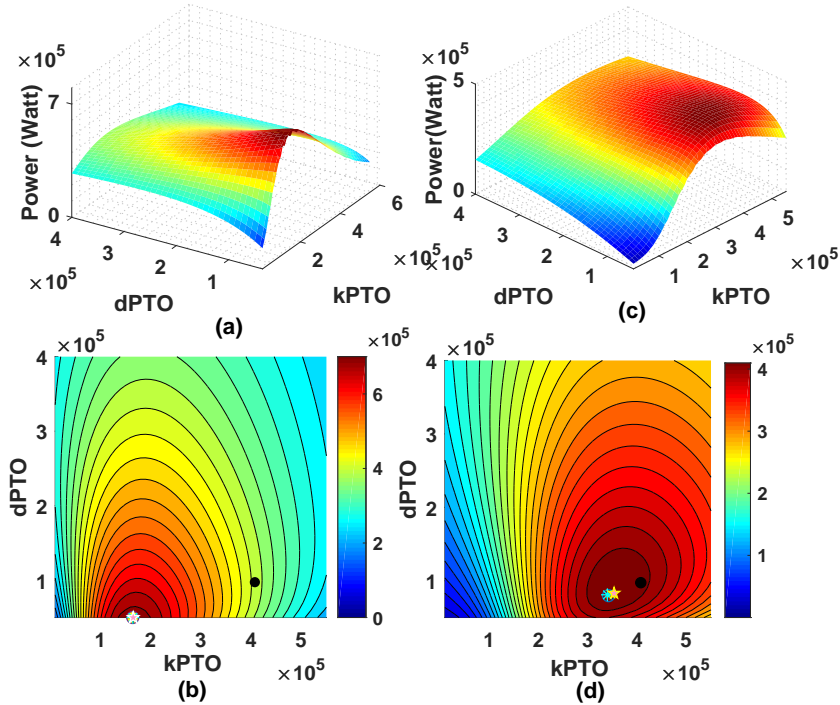}
\caption{Power landscape analysis of both real wave scenarios ((a,b) Perth, (c,d) Sydney) for the best discovered 4-buoy layouts. The spring-damping PTO configuration step size is 10000. The black circle shows the manufacturer's PTO defaults for the predominant wave frequency and the star, cross, circle, and Pentagon markers present the k and dPTO settings of the best-discovered 4 buoys layout. Note that the search space for buoy positions is multi-modal~\cite{neshat2018detailed}, and that we only visualise a 2D slice of the 8D PTO optimisation space here without considering interactions with buoys' positions.} \label{fig:landscape}
\end{figure}
\begin{figure}

\includegraphics[width=0.98\textwidth]
{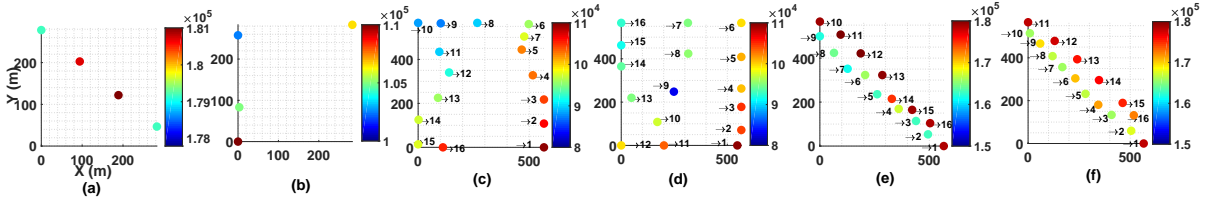}
\caption{The best-obtained 4 and 16-buoy layouts: (a) 4-buoy, Perth wave model, Power=719978.29(Watt), q-factor=1.013 by DE; (b) 4-buoy, Sydney wave model, Power=423898.52(Watt), q-factor=0.98 by DE; (c) 16-buoy, Sydney, Power= 1559605, q-factor=0.903 by SLS-NM-B1; (d) 16-buoy, Sydney, Power=1564334.59, q-factor=0.916 by SLS-NM-B2; (e) 16-buoy, Perth, Power=2739657.74, q-factor=0.966 by SLS-NM-B1; (f) 16-buoy, Perth, Power=2741489.18, q-factor=0.972 by SLS-NM-B2 (2.26\% more power than CMA-ES best layout).}\label{fig:best_solutions}
\end{figure}

\subsection{Landscape analysis}
For visualising the impact of PTO parameter optimisation, a simple experiment was done. First of all, we optimised the buoy positions for a 4-buoy layout using a manufacturer's PTOs defaults ($k=407510$ and $d=97412$) for all converters for both the Perth and Sydney test sites. The black circle in Figure~\ref{fig:landscape} marks this default PTO configuration. The energy produced by this layout is  $402$~kW and $703$~kW, respectively, for the Sydney and Perth wave climates. Next, this obtained layout is evaluated where the buoy positions are fixed and we grid-sample the energy produced when all four buoys are assigned the same PTO parameters. This process produces the contoured backgrounds shown in Figure~\ref{fig:landscape}. Finally, we optimise the PTO parameters for each buoy independently and plot a marker for each of the four buoys. These markers are roughly, but not completely, coincident with the peak in the background power landscape produced by optimising buoys' PTO parameters in unison. These markers are also at a different point to that produced by the default setting. The best energy produced after optimisation has improved to $420$~kW and $720$~kW respectively for Sydney and Perth. Another observation from Figure~\ref{fig:landscape} is that the best PTO configurations of the 4-buoy layouts are relatively alike in both wave scenarios.
\begin{figure*}[tb]
\includegraphics[ width=0.97\textwidth]
{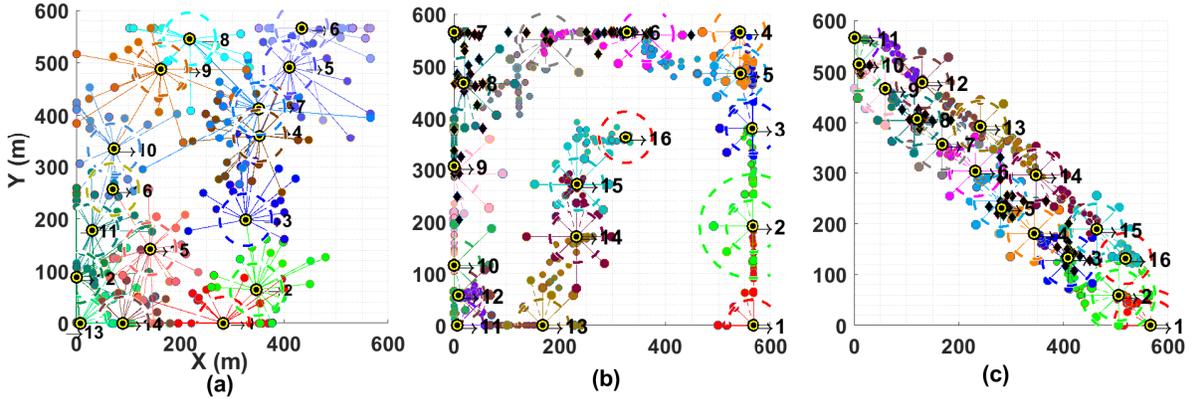}
\caption{Three illustrations of the local search process for the placement of 16 buoys using LS-NM (part (a)) and SLS-NM-B2 (parts (b) and (c)). Small yellow circles represent the final buoy positions. The coloured radial lines represent the neighbourhood sampling process. The black diamonds in parts (b) and (c) represent the positions sampled  by the backtracking algorithm. The internal circles show the safety distance and the external ones demonstrate the local search space.  Part (a) (Power=1525780W, q-factor=0.89) and (b) (Power= 1562138W, q-factor=0.91), optimise for the Sydney wave model; and part (c)(Power=2741489W, q-factor=0.972) is for Perth.}
\label{fig:three_solutions}
\end{figure*}
\begin{figure}[tb]
\centering
\includegraphics[ width=0.8\textwidth]
{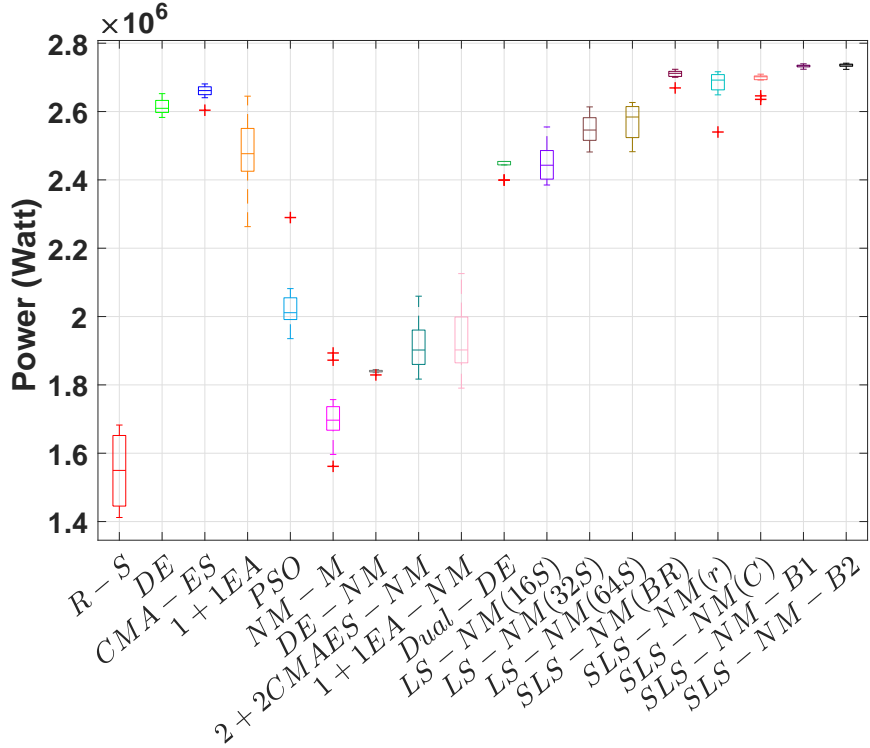}
\caption{The comparison of the proposed algorithms performances for 16-buoy layout in Perth wave model. The optimisation results present the best solution per experiment. (10 independent runs per each method)}\label{fig:boxplot_16_Perth}
\end{figure}
\subsection{Layout evaluations}
In order to evaluate the effectiveness the proposed algorithms in Sections~\ref{sec:allatonce}, \ref{sec:alternate}, and~\ref{sec:hybrid}, we performed a systematic comparison of the best layouts produced by each in two different real wave scenarios (Perth and Sydney), and for two different numbers of buoys ($N=4$ and $N=16$). Ten runs were performed for each optimisation method and the best solutions were collected for each. 

\setlength{\tabcolsep}{2pt}
 \def\arraystretch{1.05}%
\begin{table*}[tb]
\centering

\scalebox{0.55}{

\begin{tabular}{l|l|l|l|l|l|l|l|l|l|l|l|l|l|l|l|l|l}
\hlineB{4}
\multicolumn{17}{ c }{\textbf{\begin{large}Perth wave scenario (16-buoy)\end{large}}} \\
\hlineB{2}
\textbf{Methods}& \textbf{DE}&\textbf{ CMA-ES}&\textbf{1+1EA}& \textbf{PSO}&\textbf{ NM-M}& \textbf{DE-NM} &\textbf{CMAES-NM}&\textbf{ 1+1EA-NM}&\textbf{ Dual-DE}& \textbf{LS-N$M_{16s}$}&\textbf{LS-N$M_{32s}$}&\textbf{LS-N$M_{64s}$} &\textbf{SLS-NM(BR)}&\textbf{SLS-NM(r)}&\textbf{SLS-NM(C)}&\textbf{SLS-NM-B1}&\textbf{SLS-NM-B2} \\\hlineB{2}
\texttt{\textbf{Max}} & 2652393  & 2680843 & 2644987 &2289764 &1893411 & 1845065 & 2059607 & 2125726 & 2453857 & 2554865 & 2613619 & 2626506 & 2723676 & 2716463 & 2709385 & 2739658& \textbf{2741489} \\
\hlineB{2}
\texttt{\textbf{Min}}& 2582793 & 2603920 & 2263180 &1935340 &1561609 & 1829109 & 1816940 & 1790521 & 2399372 & 2384981 & 2481663  & 2482512 & 2669097 & 2540090 & 2635628 & 2723886 & 2723470 \\
\hlineB{2}
\texttt{\textbf{Mean}}& 2613938 & 2657924 & 2476649 &2034625 &1709664 & 1839680 & 1917947 & 1930481  & 2442276 & 2449269 & 2547633 & 2570651 & 2708267 & 2677821 & 2691542 & 2733105 & \textbf{2735345} \\
\hlineB{2}
\texttt{\textbf{Median}}& 2609441 & 2661285 & 2476649 &2011311 &1696728 & 1840299 & 1902074 & 1902254 & 2453857 & 2442901 & 2545870 & 2584010 & 2711875 & 2692056 & 2701771 & 2733962 & \textbf{2736453} \\
\hlineB{2}
\texttt{\textbf{Std}}& 21601.36  & 20844.29  & 109986.19  &90666.26 & 96667.21 & 4261.50 & 76927.84 & 96648.77 & 20511.38  & 53689.15 & 40651.08 & 49948.44 & 14434.14 & 48718.95  & 24252.10 & 4426.12 & 4986.80 \\ \hlineB{4}
\hlineB{4}
\multicolumn{17}{ c }{\textbf{\begin{large}Sydney wave scenario (16-buoy)\end{large}}} \\
\hlineB{2}
\textbf{Methods}& \textbf{DE}&\textbf{ CMA-ES}&\textbf{1+1EA}&\textbf{PSO}&\textbf{ NM-M}& \textbf{DE-NM} &\textbf{CMAES-NM}&\textbf{ 1+1EA-NM}&\textbf{ Dual-DE}& \textbf{LS-N$M_{16s}$}&\textbf{LS-N$M_{32s}$}&\textbf{LS-N$M_{64s}$} &\textbf{SLS-NM(BR)}&\textbf{SLS-NM(r)}&\textbf{SLS-NM(C)}&\textbf{SLS-NM-B1}&\textbf{SLS-NM-B2} \\\hlineB{2}
\texttt{\textbf{Max}} & 1544911 & 1551852 & 1550820& 1498996& 1393383 & 1372431 & 1524002 & 1541064 & 1488451 & 1525789 & 1542636 & 1551640 & 1556956 & 1550054 & 1534157 & 1559578  & \textbf{1564334}  \\
\hlineB{2}
\texttt{\textbf{Min}}& 1525043  & 1533453 & 1461996&1396223 & 1256857 & 1363834 & 1392057  & 1414872 & 1420995 & 1507479 & 1523444     & 1518276 & 1526266 & 1489493 & 1465638 & 1546369 & 1529929 \\
\hlineB{2}
\texttt{\textbf{Mean}}& 1536324 & 1547951 & 1526867&1438377
 & 1337175 & 1367502 & 1454505 & 1467659 & 1462382 & 1514404 & 1532215 & 1535923 & 1544706 & 1525152 & 1512476 & 1553629 & \textbf{1556447} \\
\hlineB{2}
\texttt{\textbf{Median}}& 1538708  & 1549616 & 1531683&1435726 & 1338054 & 1367767 & 1441785 & 1467420 & 1465419 & 1513593 & 1528728     & 1535516 & 1548100 & 1523762 & 1518423 & 1553779 & \textbf{1558319} \\
\hlineB{2}
\texttt{\textbf{Std}}& 6559.22 & 4996.61 & 25962.37&31262 & 41794.00 & 2508.76 & 47091.11 & 32623.75 & 14999.60  & 5125.37 & 7224.27 & 12944.20 & 10965.95 & 17681.23 & 18379.27 & 3293.68 & 8931.08\\
\hlineB{4}
\end{tabular}
}
\caption{The performance comparison of various heuristics for the 16-buoy case, based on maximum, median and mean power output layout of the best solution per experiment.}
\label{table:allresults16}
\end{table*}

Figure~\ref{fig:boxplot_16_Perth} shows the box-and-whiskers plot for the power output of the best solution per run for all search heuristics, for 16-buoy layouts for the Perth wave scenario. The corresponding summary statistics are presented in Table~\ref{table:allresults16} and Table~\ref{table:allresults4} for 16 and 4 converters respectively, and we illustrate the search process for three cases in Figure~\ref{fig:three_solutions}.

It can be seen that the best mean layout performance is produced by both SLS-NM-B1 and SLS-NM-B2. Additionally, the average optimisation results of SLS-NM with various first buoy locations are also competitive. Among these, the best results arise from placing the first buoy in the bottom right corner of the search space. This results in more total power output because the farm layout this placement enables a greater number of  constructive buoy interactions. Of the standard EAs, CMA-ES performs best. Interestingly, the performance of the alternating approaches is not competitive compared with other methods. 
\begin{figure}[tb]
\centering
\includegraphics[width=0.8\textwidth]
{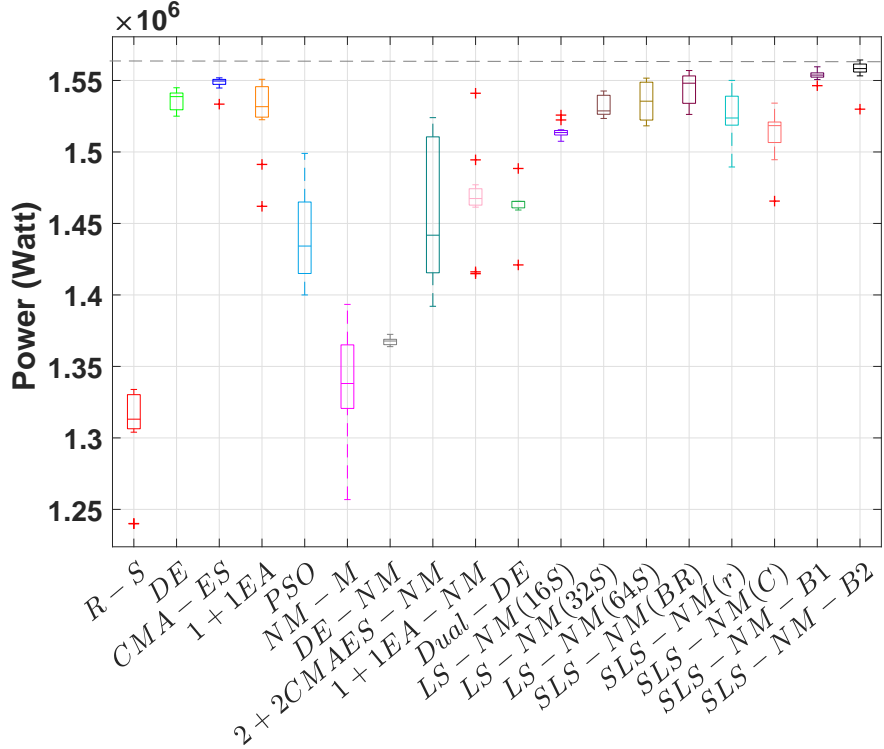}\vspace{-2mm}
\caption{The comparison of the proposed algorithms' performance for 16-buoy layouts in Sydney wave model. The optimization results present the best solution per experiment. (10 independent runs per each method)}\label{fig:boxplot_16_Sydney}
\end{figure}

Looking more closely at Table~\ref{table:allresults16}, in both wave scenarios, the SLS-NM-B2 method significantly outperforms all but the SLS-NM-B1 method using the Wilcoxon rank-sum test ($p<0.01$). The SLS-NM performs better than CMA-ES for the Perth wave model, but is no better than CMA-ES or DE for the, more challenging, Sydney scenario. This can be seen in the box-plots for the Sydney scenario shown in Figure~\ref{fig:boxplot_16_Sydney}. As a last observation, there appears to be some positive impact from increasing the number of samples in the LS-NM heuristic from 32 samples to 64. 
\begin{figure*}[tb]
\includegraphics[width=0.95\textwidth]
{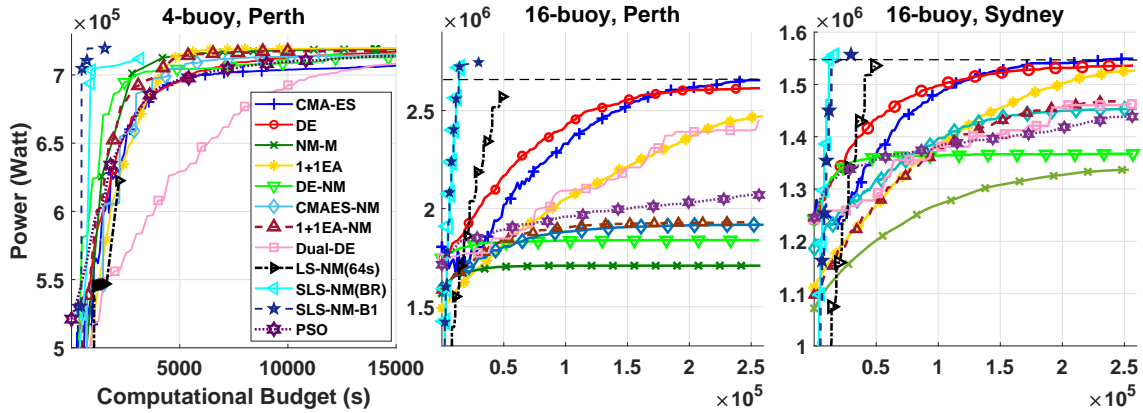}
\caption{The convergence rate comparison for all proposed algorithms in both real wave scenarios(mean best layouts per generation). Both SLS-NM(BR) and SLS-NM(B1) methods are able to place and optimise the position and PTO configurations of 4 and 16-buoy layouts faster than other proposed approaches. The horizontal dashed lines show the improvement rate difference of both SLS-NM(BR) and SLS-NM(B1) with CMA-ES.} \label{fig:all_convergence}
\end{figure*}

Figure~\ref{fig:all_convergence} shows the convergence of average fitness of the best layout over time for all of the heuristics. Part (a) shows this convergence for N =4 for the Perth model, part (b) is for N = 16 for Perth, and part (c) is for N = 16 for Sydney.

In all configurations, SLS-NM-B converges very fast and still outperforms the other methods. To sum up, the experimental results in Table~\ref{table:allresults16} and Figure~\ref{fig:all_convergence} reveal that SLS-NM-B succeeds in attaining higher absorbed power as well as faster convergence speed. A second important remark about Figure~\ref{fig:all_convergence} is that the alternating optimisation methods perform worse than the standard EAs, where both positions and PTO settings are mixed as an all-in-one problem.
One possible path to improving these alternating methods in the future could be to shift some of the budget for PTO optimisation to positional optimisation, which appears to be more challenging. 
\begin{figure}[tb]
\centering
\includegraphics[ width=\textwidth]{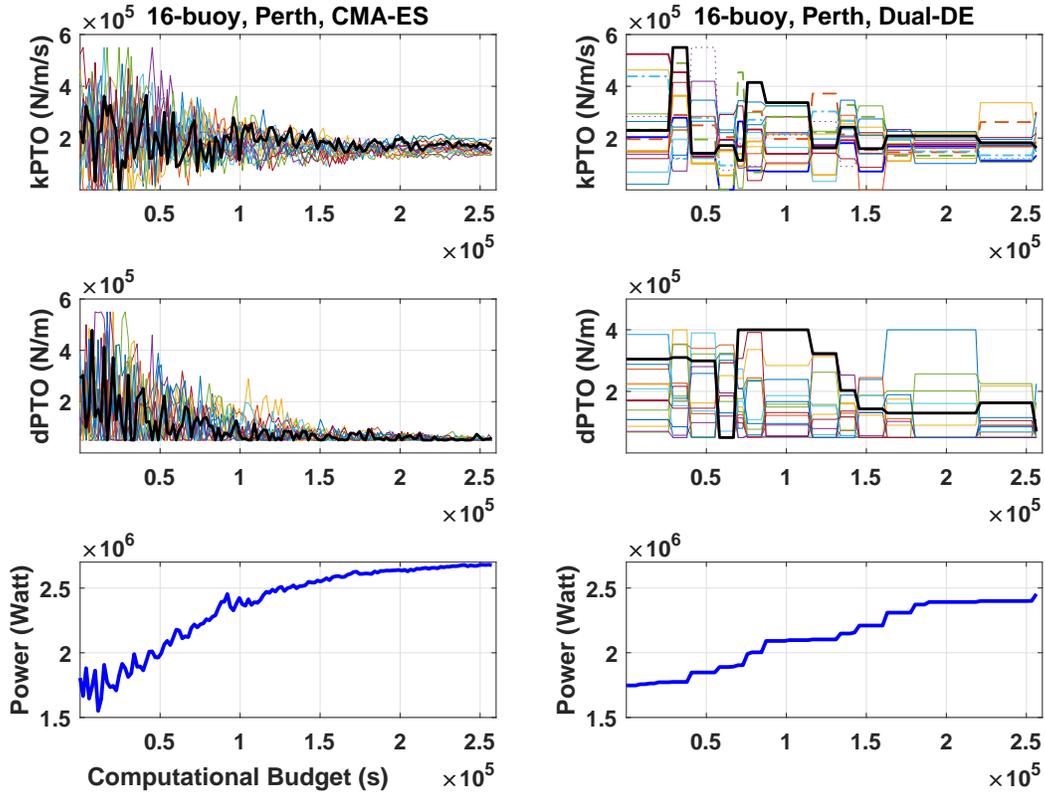}
\caption{The convergence of spring-damping PTOs of 16 buoys by CMA-ES (All-in-one) and Dual-DE (alternating style) methods in Perth wave scenario. The black line shows the $16^{th}$buoy PTO settings.}\label{fig:all_convergence_PTO}
\end{figure}
\begin{figure}
\centering
\includegraphics[ width=0.8\textwidth]
{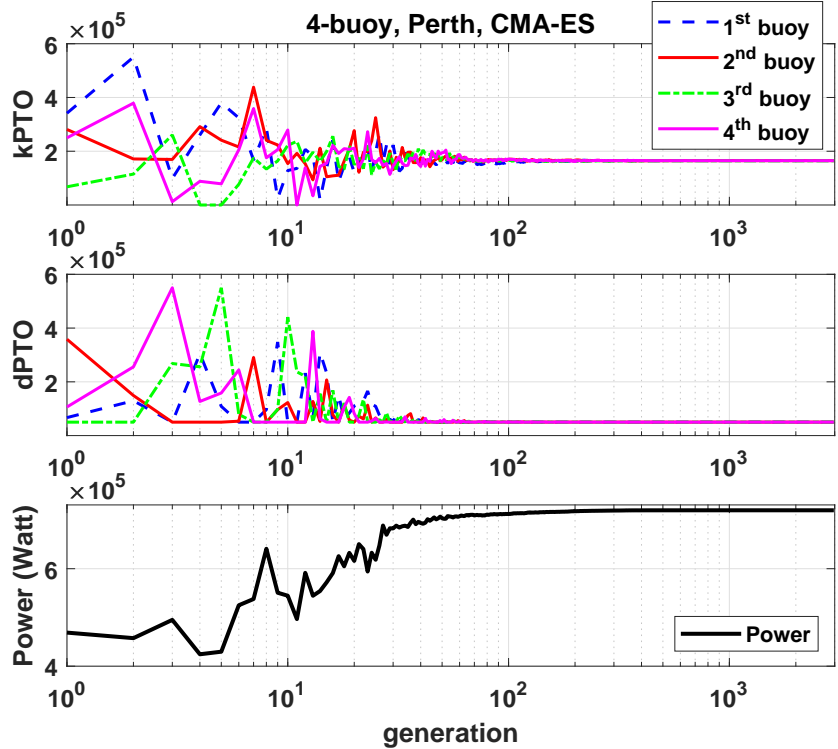}
\caption{ The convergence of spring-damping PTOs of 4 buoys by CMA-ES (All-in-one) in Perth wave scenario. } \label{fig:PTO_convergence_4}
\end{figure}

Figure~\ref{fig:all_convergence_PTO} tracks the convergence of just the PTO parameters for each buoy during a run for CMA-ES (graphs on the left) and Dual-DE optimisation (graphs on the right). It can be seen that both methods are able to optimise power output over time and the phased nature of the search in Dual-DE is visible in the graphs of the parameter values. It can also be observed that the parameter values for each buoy change non-monotonically as the best PTO settings interact with buoy positions over the course of optimisation. Meanwhile, Figure \ref{fig:PTO_convergence_4} shows the PTO optimization process of 4-buoy layouts by CMA-ES method. 

Figure~\ref{fig:best_solutions} presents the most productive 4 and 16-buoy layouts attained from all the runs in the two scenarios. The best 16-buoy layouts are built by SLS-NM-B2 from the x-axis upwards with buoys labelled, in the figure, by order of placement. In all layouts, the first buoy is placed at the bottom right.

The best 4-buoy layout of the Perth wave model slopes diagonally upwards from right to left. This layout was found by DE.  
For 16-buoys, the best SLS-NM-B2 configuration produces a maximum power output that is 2.26\% higher than the best CMA-ES configuration. Another observation is that the layouts for Sydney place buoys far from each other. This is likely to be due to the fact that the more diverse wave directions in Sydney make it harder to consistently exploit constructive interactions from having buoys in closer proximity.  

\subsection{Hydrodynamic interpretation}
\begin{figure}[tb]
\centering
\includegraphics[width=0.8\textwidth]{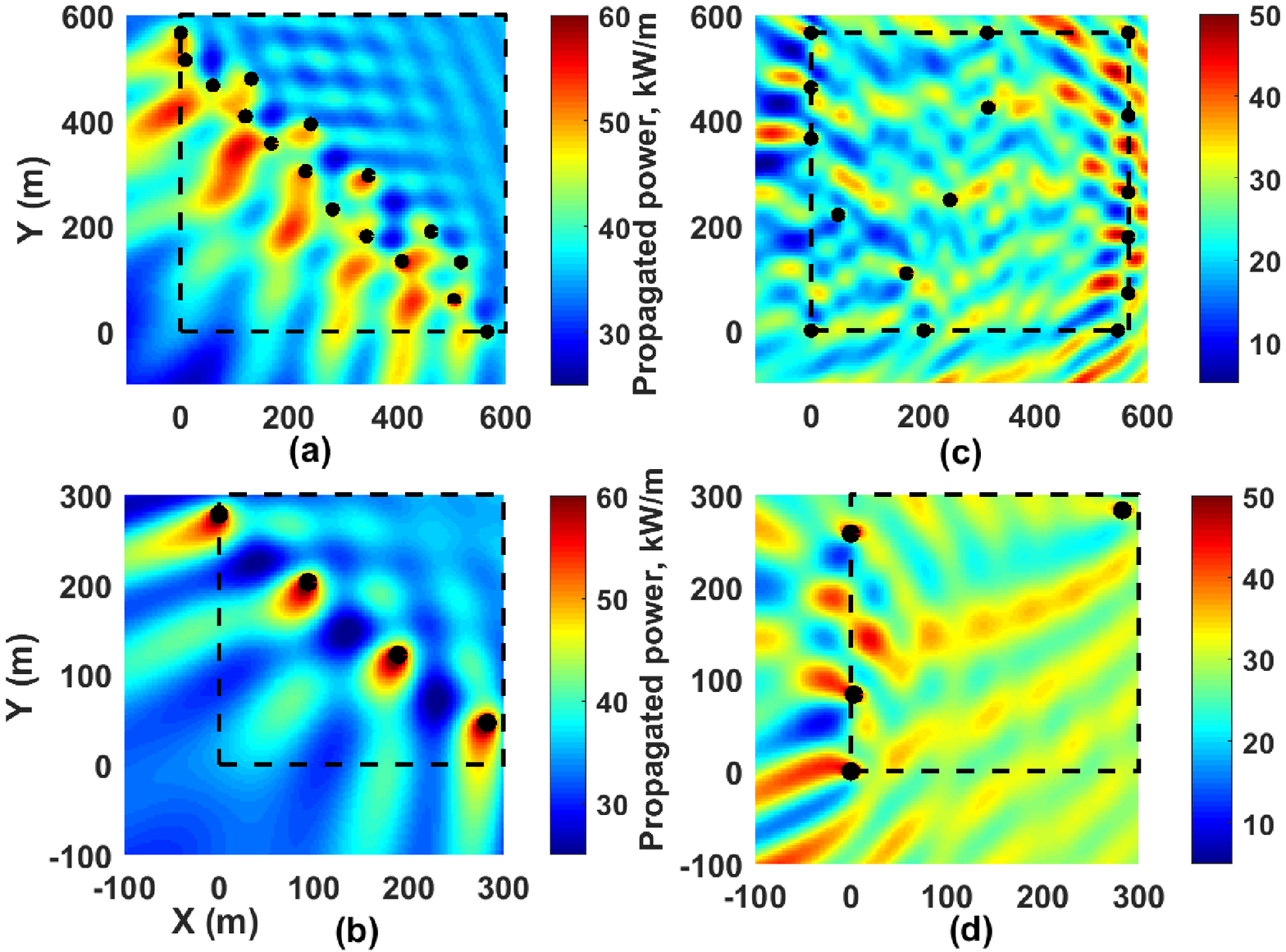}
\caption{The wave power around the best-founded 4 and 16-buoy layouts by SLS-NM-B2; (a) 16 buoys, Perth wave scenario; (b) 4 buoys, Perth; (c) 16 buoys, Sydney, and (d) 4 buoys, Sydney wave scenario. Black circles and squares show the  buoys placement and the search space.}\label{fig:wave_interpolation}
\end{figure}
 
Figure~\ref{fig:wave_interpolation} demonstrates how the ocean wave power propagates through the farm for each best-discovered solutions (4 and 16 buoy layouts) for the Sydney and Perth sites. These landscapes model interactions at the single dominant wave direction and frequency.

The wave resource at the Sydney and Perth sites is 30 and 35~kW/m, respectively. While these waves propagate through the farm, the wave field is modified by the buoys and we can see that the wave energy across the farm varies between 10 and 60~kW/m. It can be seen that, in both sites, the best layout succeeds in extracting much of the energy from the surrounding environment and, in the case of Perth, the impact of extraction extends far out to sea beyond the farm. The red areas near buoys are produced by interactions of buoys with their local environment. It should be noted that, though these areas might appear to be good candidate positions for further buoy placements, destructive interference with other buoys would produce sub-optimal results from such a placement. Another observation is that at both sites at least one row of buoys is perpendicular to the dominant wave direction (232.5 deg for the Perth site, and 172.5 deg for the Sydney site). This indicates that this wave direction can inform the initialisation of buoy positions in optimising wave farm settings.

\section{Conclusions}\label{sec:conc}
In this paper, we have described, evaluated, and systematically compared twelve different heuristic methods for optimising layout and PTO parameters for wave energy converter arrays. This study included four alternating hybrid algorithms and three new methods that are specialised to this domain. The results in this study indicate that the search problem is challenging, with buoys inducing changes in the local  power landscape and hydro-dynamic interactions occurring  between buoys.  The PTO optimisation results, also, indicate at least some interaction between buoy placement and optimal PTO settings for each buoy. Moreover, the hydrodynamic modelling required for larger buoy layouts is expensive, which constrains optimisation to take place with a limited number of evaluations. 

The best performing method is a new hybrid of a symmetric local search combined with Nelder-Mead search and a backtracking strategy. In our experiments, this method out-performed other state-of-the-art algorithms, for 16-buoy layouts, in terms of power production and in terms of speed-of-convergence. 
Future work can further improve the fidelity of the environment including considering a mix of buoy designs, tethering configurations, farm-boundaries and sea-floor shapes. These additional factors also create a more complex cost landscape, which opens the way for multi-objective optimisation. 

\vspace{3mm}
Our code, layouts, and auxiliary material are publicly available: \url{https://cs.adelaide.edu.au/~optlog/research/energy.php}
\vspace{3mm}
 \begin{table*}
\centering

\scalebox{0.58}{
\begin{tabular}{l|l|l|l|l|l|l|l|l|l|l|l|l|l|l|l|l}
\hlineB{4}
\multicolumn{16}{ c }{\textbf{\begin{large}Perth wave scenario (4-buoy)\end{large}}} \\
\hlineB{2}
\textbf{Methods}& \textbf{DE}&\textbf{ CMA-ES}&\textbf{1+1EA}&\textbf{PSO} &\textbf{ NM-M}& \textbf{DE-NM} &\textbf{CMAES-NM}&\textbf{ 1+1EA-NM}&\textbf{ Dual-DE}& \textbf{LS-N$M_{16s}$}&\textbf{LS-N$M_{32s}$}&\textbf{LS-N$M_{64s}$} &\textbf{SLS-NM(BR)}&\textbf{SLS-NM(r)}&\textbf{SLS-NM(C)}&\textbf{SLS-NM-B1} \\\hlineB{2}
\texttt{\textbf{Max}} & 719978 & 719879 & 719851&719913 & 719845 & 718321 & 718418 & 719049 & 719915 & 629667 & 633448 & 635676 & 713573  & 714041  & 703908 & 719663 \\ \hlineB{2}
\texttt{\textbf{Min}}& 719878 & 708731 & 708731 &708445 &708690 & 713598 & 706583 & 717363 & 719851 & 546821 & 600825  & 615328  & 710449 & 694667 & 701964 & 719143 \\ \hlineB{2}
\texttt{\textbf{Mean}}  & 719921& 718005 & 718491&715730  & 718914 & 717041 & 715364 & 718500  & 719882 & 599239 & 617694  & 622393 & 711976 & 704714 & 702821 & 719495 \\ \hlineB{2}
\texttt{\textbf{Median}} & 719914  & 719851 & 719850& 719107 & 719844 & 717380 & 716988 & 718653 & 719879 & 599921 & 617716 & 621512 & 711877 & 705196 & 702835 & 719554 \\ \hlineB{2}
\texttt{\textbf{Std}}  & 27.78 & 4331.96 & 3170.29& 5078.80 & 3219.83 & 1509.80 & 3925.23 & 478.99 & 28.92 & 24069.76 & 9739.71 & 5585.69  & 835.78 & 6707.32 & 563.52 & 172.24 \\ \hlineB{4}

\multicolumn{16}{ c }{\textbf{\begin{large}Sydney wave scenario (4-buoy)\end{large}}} \\
\hlineB{2}
\textbf{Methods}& \textbf{DE}&\textbf{ CMA-ES}&\textbf{1+1EA}&\textbf{PSO} &\textbf{ NM-M}& \textbf{DE-NM} &\textbf{CMAES-NM}&\textbf{ 1+1EA-NM}&\textbf{ Dual-DE}& \textbf{LS-N$M_{16s}$}&\textbf{LS-N$M_{32s}$}&\textbf{LS-N$M_{64s}$} &\textbf{SLS-NM(BR)}&\textbf{SLS-NM(r)}&\textbf{SLS-NM(C)}&\textbf{SLS-NM-B1} \\\hlineB{2}
\texttt{\textbf{Max}}  & 423898 & 423878 & 423847&423872 & 423806 & 423628 & 423485 & 423775 & 423899 & 419504 & 420549 & 420850 & 422619 & 422906 & 422878 & 422866 \\ \hlineB{2}
\texttt{\textbf{Min}} & 423489 & 422046 & 422784&420883  & 423392 & 423255 & 422464 & 423397 & 423789 & 386137 & 415848 & 413949 & 420667 & 401907 & 420125 & 420724 \\\hlineB{2}
\texttt{\textbf{Mean}}  & 423767 & 423516  & 423579&423218  & 423703 & 423406 & 423006 & 423602 & 423844 & 411155 & 418305 & 418210 & 421665 & 414943 & 421368 & 422335  \\\hlineB{2}
\texttt{\textbf{Median}} & 423808 & 423646 & 423636&423564  & 423710 & 423352 & 422988 & 423625 & 423840 & 415909 & 418262 & 418607 & 421798 & 416878 & 421421 & 422660 \\\hlineB{2}
\texttt{\textbf{Std}}  & 140.96  & 492.19 & 285.52&859.18  & 119.82 & 125.47 & 294.73 & 132.68 & 37.9 & 11436.02 & 1523.88 & 2104.40 & 624.80 & 7100.91 & 771.31 & 638.80 \\ \hlineB{4}
\end{tabular}
}
\caption{The performance comparison of various heuristics for the 4-buoy case, based on maximum, median and mean power output layout of the best solution per experiment (Std = standard deviation). In SLS-NM, the first buoy location in the search space is investigated and three options are evaluated including: Bottom right (BR), Bottom Center (C) and random (r).  }

\label{table:allresults4}
\end{table*}
\section*{Acknowledgements}
We would like to offer our special thanks to Dr. Andrew Sutton and Sagari Vatchavayi (University of Minnesota Duluth) for their valuable and constructive suggestions. Our work was supported by the Australian Research Council project DE160100850.

\bibliographystyle{unsrt}  

\bibliography{references}





\end{document}